\documentclass[sigconf, nonacm]{acmart}
\usepackage{multirow}
\usepackage{caption}
\usepackage{subfigure}
\usepackage{makecell}
\usepackage[ruled]{algorithm2e}
\usepackage{color}
\usepackage{threeparttable}
\usepackage{enumitem}

\AtBeginDocument{%
  \providecommand\BibTeX{{%
    \normalfont B\kern-0.5em{\scshape i\kern-0.25em b}\kern-0.8em\TeX}}}

\setcopyright{acmcopyright}
\copyrightyear{2018}
\acmYear{2018}
\acmDOI{XXXXXXX.XXXXXXX}

\acmConference[Conference acronym 'XX]{Make sure to enter the correct
  conference title from your rights confirmation emai}{June 03--05,
  2018}{Woodstock, NY}
\acmPrice{15.00}
\acmISBN{978-1-4503-XXXX-X/18/06}

\begin{document}


\title{Self-Supervised Text Erasing with Controllable Image Synthesis}

\author{Gangwei Jiang}
\affiliation{
    \country{University of Science and Technology of China}
}
\email{gwjiang@mail.ustc.edu.cn}

\author{Shiyao Wang}
\affiliation{
    \country{Alibaba Group}
}
\email{shiyao.wsy@alibaba-inc.com}

\author{Tiezheng Ge}
\affiliation{
    \country{Alibaba Group}
}
\email{tiezheng.gtz@alibaba-inc.com}

\author{Yuning Jiang}
\affiliation{
    \country{Alibaba Group}
}
\email{mengzhu.jyn@alibaba-inc.com}

\author{Ying Wei}
\affiliation{
    \country{City University of Hong Kong}
}
\email{yingwei@cityu.edu.hk}

\author{Defu Lian}
\authornote{Corresponding author}
\affiliation{
    \country{University of Science and Technology of China}
}
\email{liandefu@ustc.edu.cn}
\renewcommand{\shortauthors}{Jiang et al.}

\begin{abstract}
    Recent efforts on scene text erasing have shown promising results. However, existing methods require rich yet costly label annotations to obtain robust models, which limits the use for practical applications. To this end, we study an unsupervised scenario by proposing a novel \textbf{S}elf-supervised \textbf{T}ext \textbf{E}rasing (STE) framework that jointly learns to synthesize training images with erasure ground-truth and accurately erase texts in the real world. We first design a style-aware image synthesis function to generate synthetic images with diverse styled texts based on two synthetic mechanisms. To bridge the text style gap between the synthetic and real-world data, a policy network is constructed to control the synthetic mechanisms by picking style parameters with the guidance of two specifically designed rewards. The synthetic training images with erasure ground-truth are then fed to train a coarse-to-fine erasing network. To produce better erasing outputs, a triplet erasure loss is designed to enforce the refinement stage to recover background textures. Moreover, we provide a new dataset (called \textit{PosterErase}), which contains 60K high-resolution posters with texts and is more challenging for the text erasing task. The proposed method has been extensively evaluated with both PosterErase and the widely-used SCUT-Enstext dataset. Notably, on PosterErase, our unsupervised method achieves 5.07 in terms of FID, with a relative performance of 20.9\% over existing supervised baselines. 

\end{abstract}




\maketitle

\section{Introduction}
    
Text erasing has attracted increasing interest because of its wide range of applications such as privacy protection~\cite{inai2014selective}, image/video editing~\cite{wu2019editing}, and image restoration~\cite{8270072}. It aims to erase text on the stroke level by filling it with a semantically plausible background~\cite{nakamura2017scene,tursun2019mtrnet}. Most previous works ~\cite{8270072, zhang2019ensnet, tursun2019mtrnet, tursun2020mtrnet++, liu2020erasenet} focus on designing a high-quality network to remove the text from natural images. For example, EnsNet~\cite{zhang2019ensnet} is the first end-to-end framework to remove text at a whole image level. MTRNet~\cite{tursun2019mtrnet} utilizes an auxiliary mask to improve the text detection branch. EraseNet~\cite{liu2020erasenet} presents a two-stage coarse-to-fine network with an additional segmentation head, which achieves state-of-the-art performance. Although they have obtained remarkable improvements, their erasing quality strongly relies on a great amount of annotated data, which requires significant economic and labor costs. 

\begin{figure}[t]
  \centering
   \includegraphics[width=1\columnwidth]{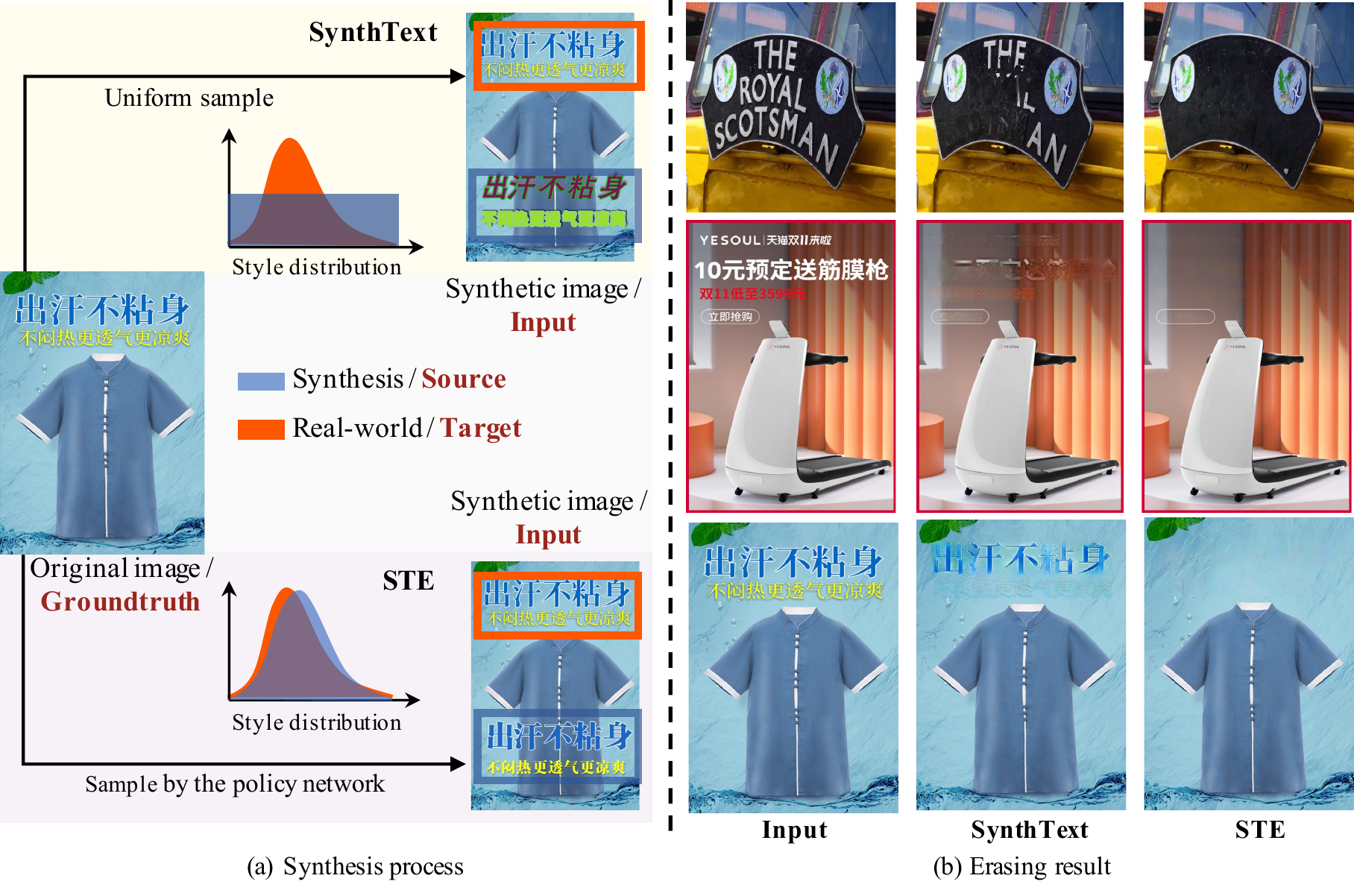}
  \caption{While the SynthText~\cite{gupta2016synthetic} is synthesized by uniform sampling from a pre-defined space, STE learns the style distribution of the real-world data by a policy network. Several erasing results on real-world data are shown on the right.
  }
  \label{task_des}
   \vspace{-0.8em}
\end{figure}

One important method to mitigate this issue is to use synthetic data. Gupta \textit{et al.}~\cite{gupta2016synthetic} used an offline generation mechanism to synthesize training image pairs. As shown in Figure \ref{task_des} (a), text styles are randomly drawn from a pre-defined style space. Then an image without the synthetic text will be regarded as the erasure ground-truth of the image blending the synthetic text.
However, remarkable distribution divergence between synthetic text style and existing text style in images leads to suboptimal results, as shown in Figure~\ref{task_des} (b).
The situation becomes even worse when being adopted for high-resolution and complex poster images (see the last two rows of Figure \ref{task_des} (b)). 
To tackle this issue, two mainstream approaches are proposed. One is to directly generate synthetic images as realistic as real-world images based on the generative adversarial networks~\cite{zhu2017unpaired, hoffman2018cycada, murez2018image}. The other is to minimize representation discrepancy between synthetic images and real-world images for style alignment~\cite{zoph2016neural,brock2018large,zhang2019bridging}. However, there are many and diverse text styles in images for erasing, making it challenging to learn a GAN network or directly minimize representation discrepancy~\cite{guan2021scale}. 

To address these issues, in this paper, we propose Self-Supervised Text Erasing (STE) framework, which consists of two modules: image synthesis and text erasing. \textbf{In the synthesis module}, in addition to generating from a customize synthesis space as in previous work, we leverage the Maximally Stable Extremal Regions (MSER) method~\cite{matas2004robust} to extract and replicate text regions from real-world images. The above \textit{customization} and \textit{replication} can provide sufficient variety to approximate the original text style.
Then, considering the text style gap between the synthetic and real-world data, a policy network is constructed to control the synthesis function under the guidance of two well-designed rewards. The rewards are calculated in terms of realistic and difficulties of the current selected style, encouraging the synthesis function to provide samples matching the target distribution while keeping the diversity.
\textbf{In the erasing module}, we use a coarse-to-fine generative model to erase the text and fill missing pixels with appropriate textures. However, the current refinement network is incapable of accurately recovering detailed information when the text or background is complex. So we propose a triplet erasure loss (TEL) to solve the issue of blurry results. 
The TEL explicitly enforces the refined results away from the coarse outputs while keeping close to the ground truth. It effectively boosts the refinement network to generate more texture details and semantics than the coarse network. The synthesis module and erasing module are optimized alternately in the training process. 
Last but not the least, we collect 60K high-resolution poster images from the e-commerce platform to embrace more challenging scenarios for  text erasing.
On the whole, our STE method gains significant improvements compared to the supervised methods in both poster and scene text.

The contributions are summarized as follows:
\begin{itemize}[leftmargin=5.5mm]
    \item We propose a novel framework (STE) for text erasing, which incorporates a synthesis function and a policy network that can produce both unbiased and diverse synthetic data. The controllable synthesis module ensures the stability and unbiased training, and effectively promotes the performance of the erasing model.
    \item The triplet erasure loss is presented to enforce the refinement network to generate more detailed and vivid content by pushing the results away from the coarse ones. It successfully improves the erasing results of text in a complex style.
    \item Additionally, the first high-resolution poster text dataset for erasing is constructed (PosterErase\footnote{will be released in the camera-ready version}), containing 60k images with text detection annotations. We conduct extensive experiments on both our dataset and the public scene text dataset. Benefiting from the improvement of data synthesis and erasing model, our method significantly outperforms all other models.   
    
\end{itemize}


\section{Related Work}

{\bf Text Erasing:} Early text erasing frameworks~\cite{khodadadi2012text, wagh2015text} usually have two stages based on traditional text detection and image inpainting techniques while being limited to easy and single-color text. Nakamura \textit{et al.}~\cite{8270072} initially designed a deep encoder-decoder network to erase text patch by patch, which makes a big success on account of the outstanding ability of deep learning. Zhang \textit{et al.}~\cite{zhang2019ensnet} presented an end-to-end architecture to erase text across the full image based on the conditional generative adversarial network~\cite{mirza2014conditional}. Inspired by the pix2pix model~\cite{isola2017image}, Tursun \textit{et al.}~\cite{tursun2019mtrnet} applied text masks as auxiliary information to achieve efficient and stable training. More recently, several works ~\cite{liu2020erasenet, tursun2020mtrnet++} improve the accuracy by explicitly modeling a branch that predicts the text regions and build the state-of-the-art text erasing models.

Training the above erasing models requires high-quality labeled pairs which are expensive and inefficient~\cite{liu2020erasenet}. To address this issue, synthesizing data from non-text images~\cite{zhang2019ensnet,tursun2020mtrnet++,liu2020erasenet} has been the most common alternative method. But the above studies primarily focus on the setting where training and test data are both synthetic, thus leaving the issue of dataset shift unattended. In addition, Zdenek \textit{et al.} ~\cite{zdenek2020erasing} used a separately trained scene text detector and an inpainting network to achieve weak-supervised learning. However, the result of this two-stage training is not favorable, since the inpainting stage loses information around text pixels.

\noindent{\bf Text Image Synthesis:} Data synthesis is an economical and efficient method for data collection in deep learning, and it has been successfully applied in fields such as text detection~\cite{gupta2016synthetic}, semantic segmentation~\cite{ros2016synthia,richter2016playing}, medical analysis~\cite{frid2018synthetic}, and so on. In text image synthesis, Wang \textit{et al.}~\cite{wang2012end} and Jaderberg \textit{et al.}~\cite{jaderberg2014synthetic} first used synthetic images in text recognition tasks. Gupta \textit{et al.}~\cite{gupta2016synthetic} developed an efficient engine that inserts diverse text at semantically relevant locations to synthesize text images. In~\cite{zhan2018verisimilar}, Zhan \textit{et al.} took semantic coherence, visual attention, and adaptable text appearance into consideration achieving verisimilar text image synthesis. 

A more advanced line of work resorts to the generative adversarial network (GAN). GA-DAN~\cite{zhan2019ga} synthesizes data by modeling the representation in geometry and appearance spaces. Wu \textit{et al.}~\cite{wu2019editing} and Yang \textit{et al.}~\cite{yang2020swaptext} developed an end-to-end trainable style retention network to modify text in images. These GAN-based methods obtain higher synthesis quality; however, they suffered from high computational overhead and generate images with lots of artifacts when out of the training distribution (\textit{e.g.}, text in shadow style).

\noindent{{\bf Domain Adaptation:}} Domain adaptation aims to study the problem of domain shift. The majority of existing research falls into two categories. Methods like~\cite{long2015learning,ganin2016domain,long2017deep,saito2018maximum,zhang2019bridging,xu2019larger} align the distribution of a source and target domain in the feature space. 
The second category aligns input images of two domains at the pixel level. Zhu \textit{et al.} ~\cite{zhu2017unpaired,hoffman2018cycada} achieved the consistency of structure and semantics by adding consistency loss on the basis of GAN models. GA-DAN~\cite{zhan2019ga} adds a multi-modal spatial learning model for the shifts in both geometry and appearance spaces.
However, the performance of these alignment methods will decline when confronted with large variance~\cite{guan2021scale}. Meanwhile, the focus on the target makes them difficult to obtain the ability outside the distribution.


\begin{figure*}[t]
\centering
\includegraphics[width=2.05\columnwidth]{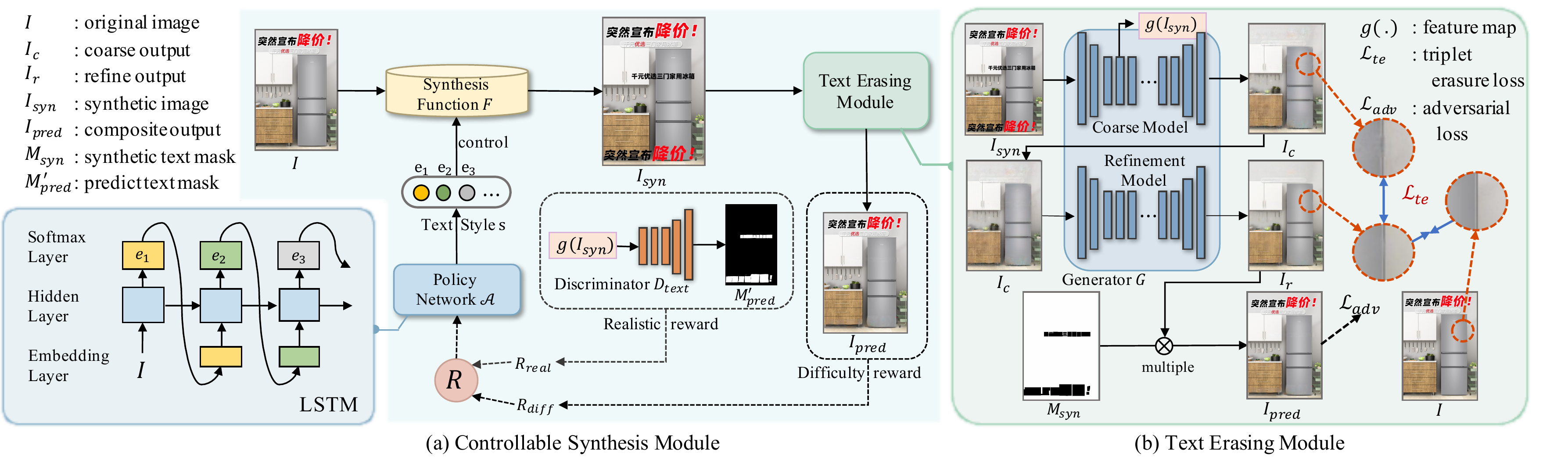}
\caption{Our Self-supervised Text Erasing framework. Part (a) presents the workflow of the controllable synthesis module. The policy network $\mathcal{A}$ is optimized with the two rewards in dashed boxes and directs the function $F$ to generate training samples. Part (b) illustrates the learning of erasing module, while the discriminator and mask prediction part is omitted. }
\label{Adg_frame}
\end{figure*}

\section{Self-supervised Text Erasing}
\label{STE}
\subsection{Overview}
\label{overview}
Our goal is to learn a text erasing network $G$, which best erases texts in a target domain (original texts in the real-world image). Given the annotated images collection $\{(I, I_{gt})\}$ where $I$ and $I_{gt}$ denote the real-world and corresponding annotated images, the previous supervised methods~\cite{zhang2019ensnet,tursun2020mtrnet++,liu2020erasenet} train $G$ by solving the following optimization problem:
\begin{equation}
    \min_{\theta}\mathcal{L}(G_\theta(I), I_{gt}),
\end{equation}
where $\theta$ are the parameters of erasing model $G$, and $\mathcal{L}$ denotes the whole erasure loss function generally consisting of adversarial loss $\mathcal{L}_{adv}$, reconstruction loss $\mathcal{L}_{rec}$, and so on. But in this paper,  we assume that we have only available to the unlabeled image collection $\{I\}$ and study an unsupervised scenario. To train $G$, we generate the synthetic image $I_{syn}$ on $I$ using a controllable synthesis module, and leverage $\{(I_{syn}, I)\}$ as the training pairs (see examples in Fig.~\ref{Adg_frame}). We learn our model $G$ with the synthetic dataset $\{(I_{syn}, I)\}$, and evaluate it on a held-out labeled test set from the target domain, which is disjoint from $\{I\}$ and is never used for training. 

Specifically, we present an overview of our self-supervised text erasing framework in Fig.~\ref{Adg_frame}. It composes of two major parts: (a) the controllable synthesis module and (b) the text erasing module. 
In the synthesis module, real-world image $I$ will be processed by synthesis function $F$ (Sec. \ref{search_space}), resulting in the synthesized image $I_{syn} \equiv F(I, s)$ with text instance in a specified style $s$. Then, $I_{syn}$ and its corresponding original $I$ can be regarded as a pair of input and ground-truth for training the text erasing model $G$:
\begin{equation}
    \min_{\theta}\mathcal{L}(G_\theta(F(I, s)), I).
\end{equation}
Moreover, to align the \textit{source} styles (synthetic text) with the \textit{target} one (original text), we use a policy network $\mathcal{A}$ to select an appropriate style $s \equiv \mathcal{A}(I)$ for function $F$, which is implemented by an LSTM ~\cite{hochreiter1997long} and optimized by the environmental feedback, including text difficult $R_{diff}$ and style realistic reward $R_{real}$ (Sec.~\ref{guided}).

In the text erasing module, our model is built upon a two-stage coarse-to-fine network called EraseNet~\cite{liu2020erasenet}, where the erasing model $G$ is construct by the coarse and refinement networks. Given the synthetic image $I_{syn}$ and a binary mask $M_{syn}$ indicating the synthesized region, the coarse network first hallucinates a rough prediction $I_c$. Then, the refinement network generates more detailed images denoted as $I_r$. And finally, the result $I_{pred}$ which used for loss calculation is composited by $I_r$ and $I_{syn}$ conditioned on $M_{syn}$. 
However, the current refinement network is incapable of generating detailed texture when the coarse network recovers indistinguishable content. So we propose a novel triplet erasure loss (TEL) $\mathcal{L}_{te}$ to ensure the refined results $I_r$ are closer to ground truth $I$ and generate more content than the $I_c$ (Sec.\ref{tel}).

The synthesis and erasing modules are jointly trained in an end-to-end manner, and thus reach better erasing learning. Specifically, the model $G$ is optimized by gradient descent while the policy network $\mathcal{A}$ is optimized with reinforcement learning. The whole loss function of our STE method can be summarized as follows:

\begin{equation}\label{erase_obj}
\begin{array}{cl}
\min_{\theta} & \mathcal{L}(G_{\theta}(F(I, \mathcal{A}_{w^*}(I))), I)\\
\text{where} &  w^* \equiv \arg\max_{w} \underset{s \sim \mathcal{A}_w(I)}{\mathbb{E}} R(I, s),\\

\end{array}
\end{equation}
$R$ denotes the weighted sum of reward function $R_{real}$ and $R_{diff}$, and $w$ are the parameters of the policy network $\mathcal{A}$.


\subsection{Style-Aware Synthesis Function}
\label{search_space}

The detailed procedure of our synthesis function $F$ is shown in Fig.~\ref{Creation}. Given an image $I$, we first utilize text detection and recognition methods~\cite{baek2019character,du2020pp} to obtain text information like position, content, blank place, and so on. Then, we generate text with a specific style $s$ and render it in a blank place. 
Different from past methods~\cite{gupta2016synthetic,zhan2018verisimilar} which only use a limited customization mechanism to generate text, we propose a simple yet powerful replication mechanism to enrich the synthesis ability. 
Details are as follows:

\begin{figure}[t]
\centering
\includegraphics[width=1.0\columnwidth]{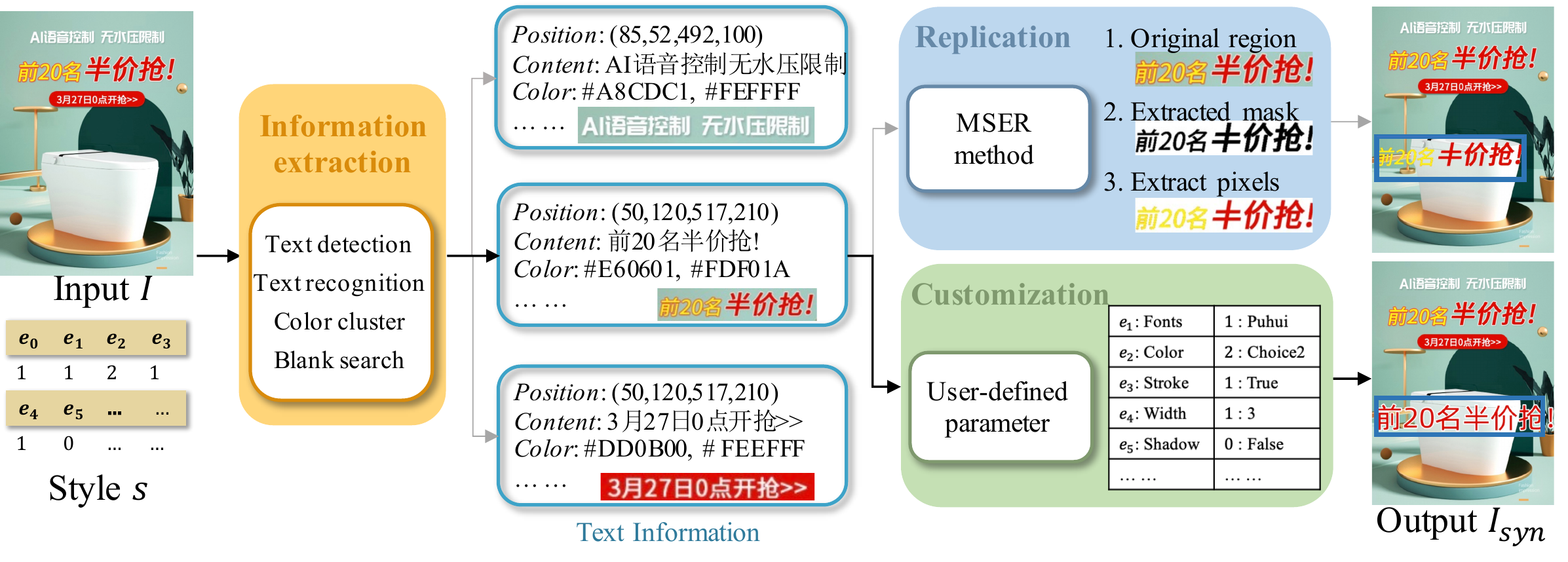}
\caption{An example procedure of the synthesis functions $F$. Specifically, $e_0$ in style $s$ denotes the choice of two mechanisms, and other elements lead the generation of the customization method. The replication method above fails to capture the pixel of the border around the text.}
\label{Creation}
\end{figure}



\noindent\textbf{Customization mechanism:} 
It was proposed in ~\cite{gupta2016synthetic}, where the text style is decomposed into multiple separated units. According to the influence on the text style, the units can be roughly divided into the following three categories:
\begin{itemize}[leftmargin=*]
    \item Appearance unit: color, font, size, Gaussian blurring, alpha blending, and Poisson blending.
    \item Geometry unit: italic, curve, and perspective transformation.
    \item Structure unit: shadow, border, and their components.
\end{itemize}
By selecting the operation in each unit, the style parameters $s$ are determined and the corresponding graphic is rendered on the image (see the output $I_{syn}$ in Fig.~\ref{Creation} as an example). 
Although the customization mechanism can create a variety of text style, they are loosely connected to the target styles and limited by the pre-defined parameter spaces. For example, The text under natural light in the first line of Fig.~\ref{task_des} (b) cannot be synthesized by it. To address the aforementioned issue, we first expand the space used in ~\cite{gupta2016synthetic} to match the huge variance in different scenarios and further propose the second replication mechanism. 

\noindent\textbf{Replication mechanism:} 
It aims to synthesize samples by copying the original text in the target distribution. To achieve this, we employ the \textbf{M}aximally \textbf{S}table \textbf{E}xtremal \textbf{R}egions (MSER) method~\cite{matas2004robust} to extract text pixels from the original area (see the top part of Fig.~\ref{Creation}). The MSER method enjoys the benefit of maintaining the target text pattern which is not limited by the space of the engine and successfully makes up for the problems incurred by the first mechanism.
Nevertheless, it may suffer from the poor extraction of style structures, \textit{e.g.}, the complex art forms like border, shadow, \textit{etc} (see Fig.~\ref{Creation} or the supplementary material for more examples). Therefore, it is important to balance the advantages and disadvantages of these two mechanisms.

Combining these two mechanisms, our synthesis function $F$ outshines in the following aspects: (a) $F$ creates samples fast and automatically; (b) the synthetic data from $F$ automatically resembles the target style by replication, and (c) $F$ is capable of synthesizing a wide range of text images by utilizing these two complementary mechanisms.
Consequently, the synthetic source data from $F$ cover the real target domain to the greatest extent and thus provides the possibility to capture the variance in the target domain.
With the synthesis function $F$ built, the synthetic samples $I_{syn}$ with original image $I$ will be fed into the erasing module as a training pair.

\subsection{Controllable Synthesis Module}
\label{guided}
Through uniformly sampling in the large space of synthesis function $F$, we can get an offline synthetic dataset \textit{SynthRC}. But the huge variance of it makes the naive training or previous synthetic-to-realistic methods~\cite{zoph2016neural,brock2018large,zhang2019bridging,zhu2017unpaired, hoffman2018cycada, murez2018image} harder to perform well. In this regard, we design a controllable synthesis module to generate realistic and harder training data in an online manner, capturing the variance in the target domain.

\subsubsection{Search Space}
The synthesis function $F$ provides various styles, all of which constitute the search space of the policy network.
Let $s = \{e_1, e_2, ..., e_N\}$ denote a sample in the space where the $e_i$ denote the $i^{th}$ style element and $N$ is the number of elements.
When each element is assigned a value, the synthesis function will generate a unique text image (see an example in Fig.~\ref{Creation}).
In particular, we have 20 elements, including MSER, font, color, \textit{etc.}, and each has a distinct selection range. Details can be found in the supplementary material. Consequently, multiple elements are composited combinatorially, leading to an exponential explosion in the number of the search space with almost $10^5$ candidates. Furthermore, our goal is to generate styles for each image, resulting in an unacceptable search space.

\subsubsection{Style Optimization via REINFORCE}
The goal of the policy network is to find a suitable synthetic style for each image in the large search space.
It is a typical discrete optimization problem since the element selection in the synthesis function are not differentiable. 
So we formulate the selection as a reinforcement learning problem and apply the REINFORCE algorithm~\cite{williams1992simple}. Concretely, the policy network is implemented as an LSTM~\cite{hochreiter1997long}. At each synthesis process, the policy network will observe the input image $I$ as the state and predict action corresponding to a discrete parameter for each element (see the policy network part in Fig.~\ref{Adg_frame}).
Notably, we construct our policy network as an image-aware agent, which is different from automatic data augmentation using a dataset-aware agent ~\cite{cubuk2019autoaugment, cubuk2020randaugment}. 
And the objective of the policy network is to maximize the reward function as Eq.~\ref{erase_obj}, and the optimization equation is as follows:
\begin{equation}
    \nabla_{w} \underset{s \sim \mathcal{A}_w}{\mathbb{E}} R(I, s) \approx \frac{1}{M*N} \sum_{m=1}^{M} \sum_{n=1}^{N} R_{m} \nabla_{w} \log p^{n}_{m}, 
\end{equation}
where $M$ is the updating batch size, $N$ is the number of element and $p^n_m$ represents the probability of the choice of the $n^{th}$ element in style $s_m$. $R$ and $R_m$ denotes the reward function and the reward value of style $m$.
The training of the policy network is inspired by~\cite{zoph2016neural, cubuk2019autoaugment, zhang2019adversarial}, and the detailed parameter updates can be found in the supplementary materials. Furthermore, to accommodate the hierarchical relationship in the search space (\textit{e.g.}, when the border style is not selected, the element controlling the border width does not work), the policy network is actually optimized using a masking mechanism as follows:
\begin{equation}
    \nabla_{w} \underset{s \sim \mathcal{A}_w}{\mathbb{E}} R(I, s) \approx \frac{1}{M*N} \sum_{m=1}^{M} \sum_{n=1}^{N} H^{n}_{m}R_{m} \nabla_{w} \log p^{n}_{m}, 
\end{equation}
where $H_m^n$ denotes the hierarchical relationship value which which is set to 1 if the element works and 0 otherwise. Details can be found in the supplementary material.


\subsubsection{Reward Setting}
\label{reward_func}
To facilitate the learning of erasing module, the generated style $s$ is expected to meet the following requirements: (a) $s$ should be realistic and match with the target distribution; (b) $s$ should be challenging for the current model, providing informative gradient; and (c) $s$ should not be extremely difficult, \textit{e.g.}, the style that generates irregular samples is not recommended. Based on these guidelines, we designed the following rewards:

\noindent\textbf{Style realistic reward:} To capture the target distribution, we implement a text discriminator $D_{text}$ to guide the data synthesis (see Fig.~\ref{Adg_frame}). Specifically, $D_{text}$ is built to predict the domain of the text region in the synthetic image $I_{syn}$. This module takes the feature map $g(I_{syn})$ of the generator $G$ as input and is compatible with any generative model plugged in.

The key to the style of realistic discriminator design is to configure the gap between the synthetic text on the training set and the original text on the validation set collected from the target distribution. So we apply an adversarial mechanism here. In the training phase, the text discriminator $D_{text}$ aims to distinguish between the original and synthetic texts, predicting the original text as 1 and other regions as 0. On the contrary, in the reward calculation phase, the closer the prediction of the synthetic text area is to 1, the greater the reward. Then the reward is formulated as (also check $R_{real}$ branch in Fig.~\ref{Adg_frame}):
\begin{equation}
\label{reward_real}
    R_{real}(I, s) =  -\left[L_{dice}(D_{text}(g(I_{syn})), M_{syn})\right],
\end{equation}
where $L_{dice}$ is the dice loss~\cite{milletari2016v} for mask image and $M_{syn}$ is the mask of synthetic text region. The realistic reward discourages the style shifts and ensures that the data synthesized is within the same distribution of target data.

\noindent\textbf{Text difficult reward:} Inspired by the success of adversarial training in distributionally robust optimization~\cite{sinha2017certifying, heinze2017conditional, volpi2018generalizing}, we consider increasing the training loss of the erasing model with hard and diverse synthetic samples. Training by the cases with larger losses guarantees robust performance against data distributions. Meanwhile, to reduce the risk of training collapse, we refer to the approach in~\cite{mao2017least,li2020pointaugment}, and finally propose a text difficult reward as follows (also check $R_{diff}$ branch in Fig.~\ref{Adg_frame}):
\begin{equation}
\label{reward_diversity}
    R_{diff}(I, s) = -\left|1- exp\left[L - \alpha L_{mean}\right]\right|,
\end{equation}
where $L$ is the L1-distance between output image $I_{pred} = G(F(I, s))$ and original image $I$, and $L_{mean}$ is the exponential average of historical loss. During training, $\alpha > 1$ controls the difficult level of generated samples, constraining that the value of $L$ stays in a certain range, \textit{i.e.}, $\alpha L_{mean}$.

Finally, the REINFORCE is employed to update the policy network using the weighted sum of the above rewards, while $R_{real}, R_{diff}$ are normalized. In practice, we update the policy network $\mathcal{A}$ for every certain number of model iterations, which makes the time overhead caused by the guided synthesis module negligible. 
\begin{equation}
\label{total_reward}
    R =  \alpha_1 R_{real} + \alpha_2 R_{diff},
\end{equation}

\subsection{Erasing Module with Triplet Erasure Loss}
\label{tel}
In this section, we will discuss how our erasing module and the newly designed loss work. 
The model in STE aims to explore the in-variance between target and source, to build the erasing ability of real-world data. However, as the labels in the target domain are not given, they may still suffer from erasure blurring, even when adopting the coarse-to-fine generator (see in Fig.~\ref{Adg_frame}).
The reason likely lies in that the current losses do not differentiate outputs of the two stages, causing the model to learn similar representations, which leads to tiny changes or fuzzy textures in the outcomes.

Inspired by the recent progress of contrastive learning, we propose our novel triplet erasure loss (TEL) to explicitly enforce the refinement network to learn different representations from the coarse network. It can be formulated as:

\begin{equation}
    \mathcal{L}_{te} = \frac{||I_r - I||_1}{||I_r - I||_1 + ||I_r - I_{c}||_1^{\gamma}},
\end{equation}
where $I, I_c, I_r$ are the label image, coarse output, and refined output, respectively. $\gamma$ is the scaling factor that emphasizes the impact of hard samples. Intuitively, TEL ensures the refined results are closer to ground truth and generates better images than the coarse results. Minimizing TEL addresses the blurring erasure problem from two perspectives. The first is that minimizing the term $||I_r - I_{c}||_1$ directly constrains the refined image standing close to its corresponding ground truth. The second is that it guides the refinement network to take a further leap from $I_c$ and generates more detailed results by maximizing the distance between them. Consequently, the loss adds a pushing force from the "negative samples" $I_c$, and the ground truth image provides an optimal direction to pull $I_r$ close to it, making the refinement network easier to converge. Notably, these two objectives mutually improve via optimization.

The training process for the erasing module has been described in Sec~\ref{overview}. Unlike the previously supervised learning~\cite{zhang2019ensnet, liu2020erasenet}, our training data $\{(I_{syn}, I)\}$ contains both labeled synthetic text and unlabeled target text in the same image. So when calculating losses, our model is expected to focus only on the synthetic text region. Thus, we combine the output image $I_r$ and input image $I_{syn}$ conditioned on the synthetic mask $M_{syn}$ to create our final prediction $I_{pred}$:
\begin{equation}
    I_{pred} = M_{syn} * I_{syn} + (1-M_{syn}) * I_{r}.
\end{equation}
Specifically, $M_{syn}$ is generated by the synthesis function without any prediction. 

Finally, we adopt the adversarial loss $\mathcal{L}_{adv}$, reconstruction loss $\mathcal{L}_{rec}$, perceptual loss $\mathcal{L}_{perc}$, style loss $\mathcal{L}_{sty}$, mask refine loss $\mathcal{L}_{m}$, and triplet erasure loss $\mathcal{L}_{te}$ as the learning objectives. The whole loss function for erasing model can be summarized as follows:
\begin{equation}
\label{e_obj}
\begin{array}{cc}
    \mathcal{L}(I_{pred}, I) =  &  \lambda_1 \mathcal{L}_{adv} + \lambda_2 \mathcal{L}_{rec} + \lambda_3 \mathcal{L}_{perc} \\
    & + \lambda_4 \mathcal{L}_{sty} + \lambda_5 \mathcal{L}_{m} + \lambda_6 \mathcal{L}_{te},
\end{array}
\end{equation}
where $\lambda_1$, $\lambda_2$, $\lambda_3$, $\lambda_4$, $\lambda_5$, and $\lambda_6$ are trade-off hyper-parameters. 

\begin{figure*}[t]
\centering
\includegraphics[width=2.10\columnwidth]{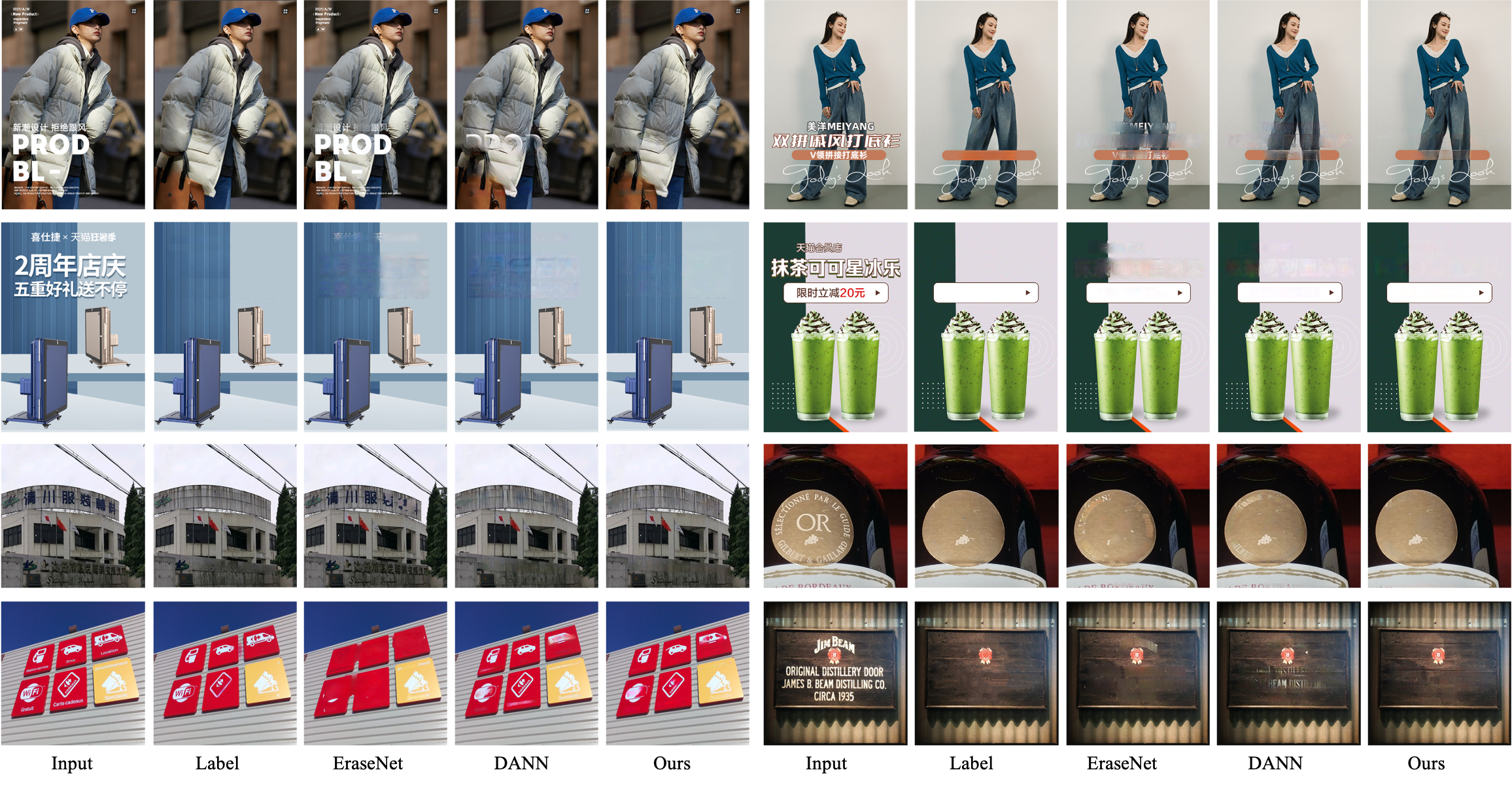}
\caption{Qualitative results for the PosterErase (the first two rows) and SCUT-Enstext (the latter two rows) based on EraseNet. More cases are included in the supplementary material. }
\label{showcase}
\end{figure*}

\section{Experiment}
In this section, we conduct experiments on PosterErase and SCUT-Enstext~\cite{liu2020erasenet} datasets to investigate the erasing quality, synthesis ability, and robustness of our self-supervised text erasing method. Unless otherwise specified, the reported results below are the performance tested on the PosterErase dataset.

\subsection{Dataset}
\label{dataset}

\textbf{SCUT-Enstext:} It is a widely used benchmark for scene text erasing collected by ~\cite{liu2020erasenet}. It contains 2,748 training images and 813 test images with their annotations. Since our method is a self-supervised framework, it can benefit from increased amounts of unlabeled text images. So we adopt 10,166 scene text images without annotations from ICDAR2019-Art dataset~\cite{chng2019icdar2019} as auxiliary information. All the images are distinct and resized to $512\times512$.

\noindent{\textbf{PosterErase:}} The dataset is collected on the e-commerce platform and consists of posters mainly with Chinese text. The dataset contains 60,000 training images and 400 test images. Each image is released with its text information, including the bounding box and text content. We also provide the annotated images for the test set, carefully processed by human experts using Adobe Photoshop (PS). Through the test dataset, we can quantitatively evaluate the quality of the model.
Each image is in the size of $750\times513$ and is resized to $768\times512$ as the model input.

\subsection{Experimental Settings \& Evaluation Metrics}
We use Adam~\cite{kingma2014adam} to train models, where $\beta_1 = 0.5$ and $\beta_2 = 0.9$. The learning rate is set to 0.0001, 0.00001, and 0.00005 in the generator, discriminator, and policy network, respectively. We set the batch size to 32. In the synthesis function, we adapt the CRAFT~\cite{baek2019character} and PaddleOCR~\cite{du2020pp} model to detect and recognize text. The policy network receives the image representations extracted by VGG-16 as input. It is implemented by a two-layer LSTM and updated every 150 batches of erasing network updating. In each update process, 100 images in the training set will be used to calculate the reward for text styles. For the construction of the erasing network, we followed the settings in the papers~\cite{liu2020erasenet}, and the experimental results are obtained from the same pre-trained initial point.

Three commonly used image generation metrics are used to evaluate the quality of the method: PSNR, SSIM and, FID. PSNR describes the error between pixels, SSIM~\cite{wang2004image} evaluates the structural similarity of the two images, and FID~\cite{heusel2017gans} compares the quality of the generated pictures from the feature level. The larger PSNR, SSIM, and the smaller FID symbolize the higher the quality of model generation. In addition, we also provide the visualization results (Fig.~\ref{showcase}) to show the model's effect directly. 

\begin{table*}[]
\caption{Comparison with baselines in PosterErase and SCUT-Enstext dataset. Higher PSNR, SSIM, and Lower FID is better.}
\label{tab:comparison_baselines}
\begin{tabular}{@{}c|c|lll|lll@{}}
\toprule
        \multicolumn{2}{l|}{\multirow{2}{*}{}}         & \multicolumn{3}{c|}{PosterErase}     & \multicolumn{3}{c}{SCUT-Enstext~\cite{liu2020erasenet}}     \\ \cmidrule(l){3-8} 
        \multicolumn{2}{l|}{}  & SSIM($\uparrow$)   & PSNR($\uparrow$)   & FID($\downarrow$)    & SSIM($\uparrow$)    & PSNR($\uparrow$)   & FID($\downarrow$)    \\ \midrule
\multirow{5}{*}{Text Erase} & Pix2pix~\cite{isola2017image}  & 0.9160  & 26.828 & 7.672  & 0.5343 & 23.865 & 17.05   \\
& EnsNet~\cite{zhang2019ensnet}         & 0.9399 & 30.196 & 7.409  & 0.8245 & 30.785 & 6.548  \\
& MtrNet++~\cite{tursun2020mtrnet++}       & 0.8326 & 26.921 & 10.11  & 0.8754 & 30.982 & 7.379    \\
& EraseNet~\cite{liu2020erasenet}       & 0.9389 & 34.034 & 6.411 & 0.8844 & 32.092 & 5.567   \\
& SceneTextErase~\cite{zdenek2020erasing} & 0.8165 & 28.990  & 11.27   & 0.8758 & 32.307 & 6.260  \\ \midrule
\multirow{5}{*}{\makecell[c]{Synthetic \\to \\Realistic}} & SynthText~\cite{gupta2016synthetic}      & 0.9363 & 32.042 & 6.599   & 0.8796  & 31.839 &  6.069  \\ 
&DANN~\cite{ganin2016domain} & 0.9433  &  35.728 & 5.625  & 0.8837  & 32.939 & 5.395 \\
&PCD~\cite{yue2019domain}    &   0.9457     & 35.601 & 5.622 &   0.8823    &   32.751     & 5.542 \\
&AFN~\cite{xu2019larger}    &   0.9463     & 36.098 & 5.269 &   0.8827    &   32.650     & 5.675 \\
&AA ~\cite{zhang2019adversarial}   & 0.9353 & 33.477 & 6.219  & 0.8812   &  32.669  &5.928 \\ \midrule
\multirow{3}{*}{Ours} & SynthRC & 0.9418 & 34.707 & 5.945   &    0.8829    & 32.765  &   5.480     \\
&STE           & 0.9548 & 37.249 & 5.070    &  0.8867& 33.198	& 4.990  \\ 
& STE+Finetune  & \textbf{0.9648}  &  \textbf{39.914}  &  \textbf{4.093} &   \textbf{0.8915} & \textbf{34.139} &  \textbf{4.318}  \\ \bottomrule
\end{tabular}
\end{table*}

\subsection{Comparison with the state-of-the-Art}
In this subsection, we show the performance of the related methods in Table~\ref{tab:comparison_baselines} and Figure ~\ref{showcase}. The baseline methods can be divided into two groups: supervised text erasing models and synthetic-to-real adaptation methods.
\begin{itemize}[leftmargin=*]
\item \textbf{Supervised text erasing models:} the first includes the various supervised/weak-supervised erasing models, including \textit{Pix2pix}~\cite{isola2017image}, \textit{EnsNet}~\cite{zhang2019ensnet}, \textit{MtrNet++}~\cite{tursun2020mtrnet++}, \textit{EraseNet}~\cite{liu2020erasenet} and \textit{SceneTextErase}~\cite{zdenek2020erasing}. They are trained on the annotated data and then tested. For the PosterErase dataset which only contains 150 training pairs, we take the well-trained model on SCUT-Enstext as initialization and achieve better results.
\item \textbf{Synthetic-to-real adaptation models:} the second type focuses on the utilization of synthetic data. \textit{SynthText} is the offline synthetic dataset proposed by ~\cite{gupta2016synthetic}, while \textit{DANN}~\cite{zoph2016neural}, \textit{PCD}~\cite{yue2019domain}, \textit{AFN}~\cite{xu2019larger}, and \textit{AA}~\cite{zhang2019adversarial} are a series of methods dealing with dataset shift. Specifically, \textit{DANN} and \textit{PCD} capture the consistency of features between domains, \textit{AFN} adopts feature-norm across domains, and \textit{AA} applies adversarial data augmentation. These methods are trained on the synthetic dataset \textit{SynthRC} and then tested on real-world images. 
\end{itemize} 
The hyper-parameters of each baseline are carefully designed to ensure the best possible results.


We compare Self-supervised Text Erasing to the above baselines on both PosterErase and SCUT-Enstext datasets. Specifically, \textit{SynthRC} is the offline synthesis method proposed in this paper which is based on replication and customization mechanism. 
The experimental results are reported in Table~\ref{tab:comparison_baselines}. The key finding is that the STE algorithm achieved excellent erasing performance without annotated data by effectively solving the inconsistency between the synthetic samples and the target data while providing diversity. Other findings include the following:

\noindent\textbf{Quantitative evaluation:} Firstly, our algorithm outperforms the state-of-the-art supervised methods in text erasing, with the improvements on FID of 20.9\% and 10.3\% in PosterErase and SCUT-Enstext. This is due to the fact that the supervised models are learned by limited labeled data, while STE provides the method for introducing richer training materials.
Second, when compared with the algorithms in the synthetic-to-realistic topic, our method still achieves significant improvements on FID of 3.8\% and 7.5\% in two datasets. By regulating the synthesis process, our method could match the variance in target from the large space of synthesis function and provide generation ability to out-of-distribution images. Thus, the erasing model can be trained stably and fast, and perform well in the target scene. Finally, we tried to evaluate our STE in a semi-supervised setup by unsupervised pretraining followed by finetuning using a small number of labeled samples. The result in the last raw of Table~\ref{tab:comparison_baselines} demonstrates that our self-supervised model with massive unlabeled data provide an effective prior for model initialization and help to yield better performing.

\noindent\textbf{Quantitative comparison:} We also present several erasing examples on both datasets in Fig.~\ref{showcase}. The erasing results of EraseNet~\cite{liu2020erasenet} and DANN~\cite{ganin2016domain} are compared to our outcomes as they have the best erasing performance in the two types of baselines. It is evident that STE generates the most visually appealing results with the fewest noticeable artifacts. More findings are as follows. First, the model from supervised training(\textit{i.e.}, EraseNet in the figure) performs poorly in detecting the text area and completing the background. This is primarily due to the limited scale of training data, which makes the model difficult in generalizing to new cases. Second, the DANN lacks the capacity to process special style texts (\textit{e.g.}, the shadow on the case in the second row of the first column and the border on the cases in the first to second rows of the second column have not been erased), but our method improves the erasing results of these images by altering the synthetic data distribution. 

\subsection{Ablation Study}
To validate the effectiveness of each component, we conduct comprehensive ablation studies in this section. The majority of the experiments were performed on the PosterErase.

\begin{figure}[t]
\centering
\includegraphics[width=1\columnwidth]{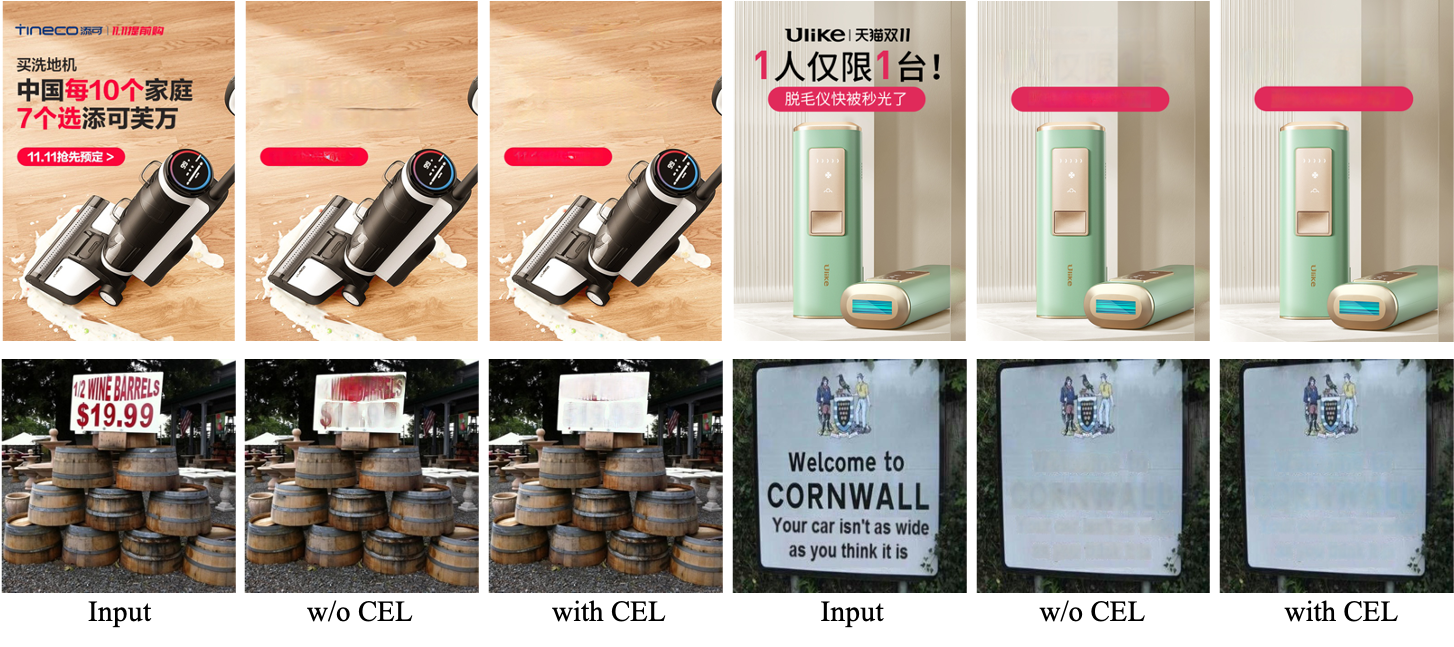}
\caption{Qualitative analysis of triplet erasure loss on both PosterErase (the first row) and SCUT-Enstext (the second row) datasets. }
\label{con_quant}
\end{figure}

\noindent\textbf{Triplet erasure loss:}
We employ triplet erasure loss $\mathcal{L}_{te}$ to help refinement network in producing clear texture. To demonstrate its superiority, we compare it with the baseline without using $\mathcal{L}_{te}$.  The numerical results in Table~\ref{constrative} show that with the help of TEL (the $\lambda$ in $\mathcal{L}_{te}$ is set to 2 to get the best performance), the FID metric drops considerably. In addition, the qualitative study in Fig.~\ref{con_quant} shows that triplet erasure loss recovers more details, which further benefits the quality of the erasing results.

\begin{table}[ht]

    \small
    \caption{Ablation study on triplet erasure loss (TEL). }
    \centering
    \label{constrative}
    \begin{tabular}{c|cc|cc}
        \toprule
        & \multicolumn{2}{c|}{PosterErase} & \multicolumn{2}{c}{SCUT-Enstext} \\ \cmidrule(r){2-3} \cmidrule(r){4-5} 
            & psnr  & fid   & psnr    & fid    \\ \midrule
    w/o $\mathcal{L}_{te}$  &    36.796  &  5.277 &    32.947	& 5.264     \\
    with $\mathcal{L}_{te}$ & 37.249 & 5.070      &  33.198	& 4.990      \\ \bottomrule
    \end{tabular}
\end{table}

\noindent\textbf{Effectiveness of style optimization components:} Table~\ref{reward function} shows the impact of each component in the optimization process of the controllable synthesis module. We study three variants: 1) using only realistic reward $R_{real}$; 2) using both two rewards $R_{real}$ and $R_{diff}$; 3) using the image as the environmental state in REINFORCE. Experiments show that the complete optimization design has gained 2.09 improvements on PSNR. We also find that $R_{real}$ helps in finding the style in target distribution, whereas $R_{diff}$ is more likely to yield complex samples, and the final reward function effectively combines the benefits of both. Here $R_{real}$ takes the greatest role in guiding the synthesis of training data. On the other hand, considering the image as a state enriches the ability of the synthesis module to model text styles, capturing the personalized demands, and hence improves the synthesis performance.

\begin{table}[ht]
    \caption{The influence of the component in the optimization of the controllable synthesis module. Image-aware indicates that the optimization takes the image as the state. 
    }
    \centering
    \label{reward function}
    \begin{threeparttable}
        \begin{tabular}{ccc|cc}
            \toprule
            $R_{match}$& $R_{diff}$ & $image-aware$ & PSNR($\uparrow$) & FID($\downarrow$)\\
            \midrule	
        & &  & 34.71 & 5.946\\
        $\checkmark$	 &  &  & 36.07 ($\uparrow$1.36) & 5.612 ($\downarrow$0.334)\\
        $\checkmark$	& $\checkmark$ &  & 36.29 ($\uparrow$1.58)& 5.483 ($\downarrow$0.463)\\
        $\checkmark$	& $\checkmark$ & $\checkmark$ & 36.80 ($\uparrow$2.09)& 5.277 ($\downarrow$0.669)\\
            \bottomrule
        \end{tabular}
    \end{threeparttable}
\end{table}





\begin{figure}[t]
\centering
\includegraphics[width=1.00\columnwidth]{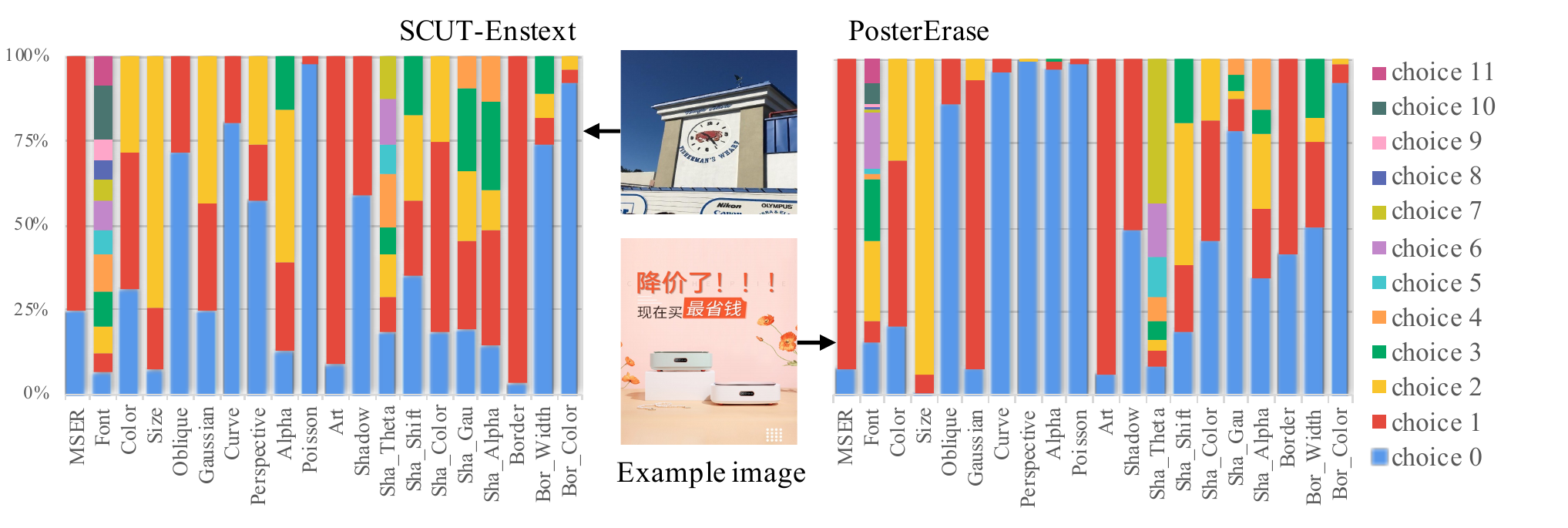}
\caption{Visualization results in the synthesis probabilities of style elements. The horizontal axis shows the elements in the search space. The vertical axis shows the proportion of each parameter in the corresponding element.}
\label{textstyle}
\end{figure}

\noindent\textbf{Text style output by controllable synthesis module:} 
A major contribution of our work is to guide the synthesis function to automatically produce realistic and harder training pairs. Therefore, we visualized the synthesis probabilities of each element in the output text style. The probability value is obtained by counting the element selection of 2,000 images under the guidance of the policy network. The results are shown in Fig.~\ref{textstyle}. The parameter choices of each element are represented by "choice0-11", where the detailed information is listed in the supplementary material. Generally, the higher the number, the deeper the degree of operation. For example, "Alpha0-3" represents 0-30\% of the background involved during alpha blending. We can see that there is a significant difference in the elements of gaussian, alpha, curve, and perspective between the two datasets, which is congruent with our common sense of natural and poster scenarios. The results reveal that our controllable synthesis module is effective in resolving the dataset shift issue.

\noindent\textbf{Robustness to different style spaces:} 
We also check the performance of the STE in different spaces of the synthesis function $F$ to confirm its effectiveness. The results are summarized in Fig.~\ref{space}. We construct four pre-defined spaces and detailed information is available in the supplemental material. As the style space rises, the variance in the source domain increase and the performance of baselines improves and then declines. Instead, our STE continually maintains efficiency and significantly outperforms them. Here, the drop of \textit{SynthRC} on space2, 3 is related to the introduction of the "Perspective" element which rarely occurs in the poster scene.

\begin{figure}[t]
\centering

\includegraphics[width=0.8\columnwidth]{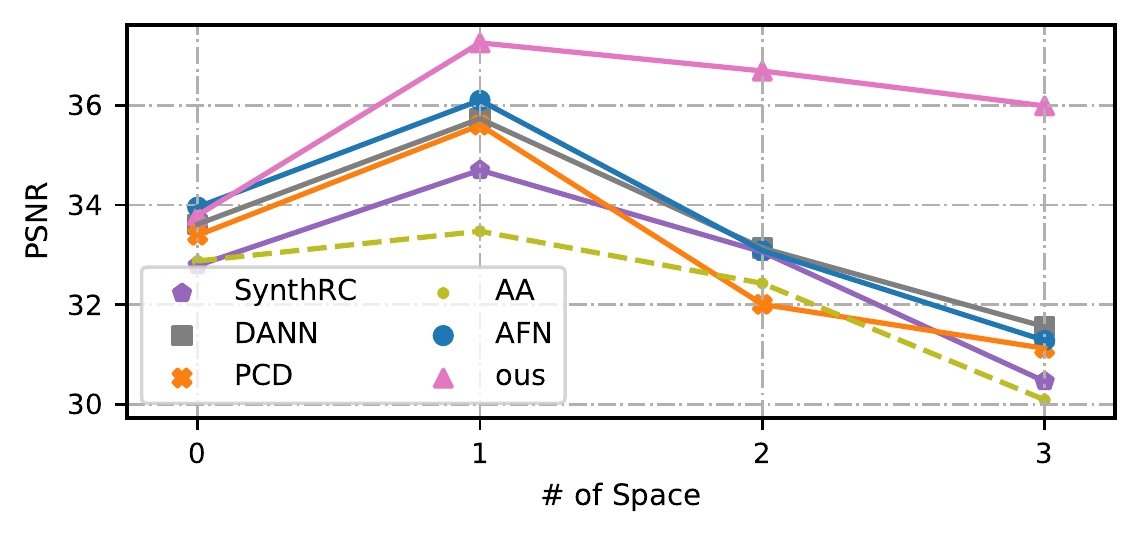}

\vspace{-0.5em}
\caption{The performance of STE and baselines on different style spaces on PosterErase. The space expands with the increase of the serial number.}
\label{space}
\end{figure}

\begin{table}[h]
    \caption{Ablation study on elements of the synthesis function. Bas., Rep. and Cus. are abbreviations of basic, replication, and customization respectively. The reported metrics is PSNR.
    }
    \centering
    \label{synthsis component}
    \begin{threeparttable}
        \begin{tabular}{ccc|c|c}
            \toprule
            Bas. & Rep. & Cus.  &  PosterErase & SCUT-Enstext\\
            \midrule
            $\checkmark$ & &   & 27.518 & 31.887\\
            $\checkmark$ &$\checkmark$  & & 31.110 & 32.841\\
            $\checkmark$ & $\checkmark$ &$\checkmark$&  34.707 & 32.765\\
            \bottomrule
        \end{tabular}
    \end{threeparttable}
\end{table}

\noindent\textbf{Analysis on the synthesis function:} We demonstrate the effectiveness of each component in the synthesis function (also see Sec.~\ref{search_space}), and the results are reported in Table~\ref{synthsis component}. By adding the replication mechanism, and the customization mechanism in the synthesis function, the performance of the trained model can raise from 27.518 to 34.707. Specifically, they are all trained without the guides of the policy network. Such a result shows that with the various text styles in the target task that can be effectively included in the synthesis function, higher performance of the model training can be achieved. Notably, replication works better in natural scenes. This is because the text changes of natural scenes are concentrated on appearance and geometry, which can be well copied by MSER. 
  
\section{Conclusion}
In this paper, we propose a novel framework, named Self-supervised Text Erasing (STE), to learn the generated training image pairs in an unsupervised fashion for the text erasing task. Specifically, our proposed STE method is composed of two newly developed modules: text synthesis and text erasing. The synthesis module is responsible for fast generating training samples as well as learning a policy network for steering the synthesis process by selecting more realistic and harder training data.  The text erasing module employs a so-called triplet erasure loss to better recover background textures.  The two modules are alternately updated in the training process. We collect 60K high-resolution poster images from the e-commerce platform to embrace more challenging scenarios for  text erasing.
Extensive experiments clearly show that our proposed unsupervised STE method outperforms existing supervised baselines, which demonstrates the effectiveness of the text synthesis and erasing strategy in our STE. 

\bibliographystyle{ACM-Reference-Format}
\bibliography{egbib}

\end{document}